\documentclass[runningheads]{llncs}
\usepackage[final,year=2026]{eccv}
\usepackage{eccvabbrv}
\usepackage[utf8]{inputenc}
\usepackage{graphicx}
\usepackage{booktabs}
\usepackage{amsmath,amssymb}
\usepackage{multirow}
\usepackage[normalem]{ulem}
\useunder{\uline}{\ul}{}
\usepackage{pifont}
\usepackage[accsupp]{axessibility}
\usepackage{hyperref}
\usepackage{bibunits}  

\makeatletter
\newcommand{\printfnsymbol}[1]{\textsuperscript{\@fnsymbol{#1}}}
\makeatother



\begin{document}

\title{Resource-Efficient RGB-Only Action Recognition for Edge Deployment}
\titlerunning{Resource-Efficient RGB-Only Action Recognition}

\author{Dongsik Yoon\thanks{Equal contribution.} \and
Jongeun Kim\printfnsymbol{1} \and
Dayeon Lee\printfnsymbol{1}}
\authorrunning{D. Yoon, J. Kim, and D. Lee}
\institute{HDC LABS, Republic of Korea}

\maketitle

\begin{bibunit}[splncs04]

\begin{abstract}
Resource-constrained assistive monitoring requires compact local video perception and an explicit understanding of how recognition reliability changes under common visual degradation. We present a compact RGB-only action-recognition network that combines an X3D-style hierarchy with temporal shift, a 3D Universal Inverted Bottleneck, Ghost pointwise convolutions, factorized depthwise operators, selective temporal adaptation, and parameter-free attention. The model attains $95.10/98.31\%$ on NTU RGB+D 60 (X-Sub/X-View) and $90.88/92.66\%$ on NTU RGB+D 120 (X-Sub/X-Set) with only $0.96$--$0.99$M parameters. To connect compact deployment with assistive use, we evaluate frozen checkpoints without retraining on the datasets' official Medical Conditions groups while retaining the original 60-/120-way decision spaces. Medical macro recall reaches $93.19/97.93\%$ on NTU60 and $91.47/91.48\%$ on NTU120. Under controlled Level-2 visual stress on NTU60 X-Sub, medical recall drops by $1.77$ points under low light, $2.23$ under motion blur, $4.87$ under crop perturbation, and $19.05$ under occlusion; among the four selected Level-2 perturbations, occlusion causes the largest observed degradation. On Jetson Orin Nano, the TensorRT FP16 engine occupies $5.30$\,MiB, demonstrating strong static compactness but not throughput superiority. The results position the model as a compact perception component for lower-rate assistive monitoring, rather than as a complete clinical or robotic system.
\end{abstract}

\section{Introduction}

Human action recognition has long been a core challenge in video understanding, and remains central as deep learning continues to advance real-world perception capabilities across a broad range of vision tasks---from efficient super-resolution and real-world image restoration~\cite{yoon2024overallnet, yoon2025srrealworld} to high-level recognition~\cite{survey1, survey2}. A growing motivation is assistive monitoring in homes, smart environments, and human-centered systems~\cite{debes2016monitoring,das2019toyota,crispim2017behave,chen2025opph}, where local video perception must run under limited compute, storage, memory, and power. RGB-only inference is attractive in this setting because the recognition backbone consumes a single RGB stream and requires no auxiliary depth, skeleton, or pose stream at inference time. Compactness alone, however, is not sufficient: assistive use also requires visibility into class-specific failure and sensitivity to common visual degradation such as low light, motion blur, occlusion, and crop errors. A system must therefore capture not only the visual content of a scene and the temporal structure of actions, but also how reliability changes under imperfect input, all under hardware budgets far tighter than those assumed by datacenter-scale training and evaluation.

As action recognition is integrated into real-world products and infrastructure, execution has shifted from datacenter GPUs to resource-constrained edge platforms such as embedded GPUs, on-device NPUs, and smart cameras. In many applications edge-side inference is a necessity rather than a design choice: it reduces latency, avoids reliance on unstable network connectivity, and mitigates privacy and bandwidth concerns by limiting raw video transmission. At the same time, moving action recognition models to edge devices introduces deployment-critical constraints that are often underrepresented in algorithm-centric evaluations. Low-power devices must operate under strict limits on memory, storage, and power or thermal budgets, frequently while sharing resources with other on-device workloads such as multi-camera decoding, detection, tracking, and system monitoring.

These constraints give rise to several engineering challenges: model graphs that are difficult to optimize within embedded inference runtimes, large serialized engines that increase storage and update costs, elevated host and device memory usage that restrict workload concurrency, and sustained power draw that can trigger thermal throttling. As a result, deployment-aware evaluation must go beyond offline accuracy and consider metrics such as parameter count, serialized engine size, peak host memory, power consumption, and inference-time device memory usage. Crucially, these axes do not always move together: a model can be extremely compact in terms of parameters and serialized size while still being limited in throughput by the way its operators map onto a particular embedded runtime. Recognizing this separation, rather than collapsing all efficiency notions into a single ``efficient'' label, is one of the framing principles of this paper.

Prior work has explored several directions to address these challenges. RGB-only action recognition has advanced through improved objectives and representation learning~\cite{tran2015, baradel2018, zhu2019, vyas2020, cheng2022}, while lightweight architectures have focused on efficient temporal modeling and mobile-friendly designs~\cite{lin2019tsm, epam}. In parallel, many recent approaches improve accuracy--efficiency trade-offs by leveraging additional modalities, either at inference or as training-time guidance~\cite{epam, liu2024multi, sv}. Although effective in benchmarks, reliance on auxiliary modalities at inference is often undesirable for real-world operation, as it requires external sensors or compute-intensive preprocessing such as pose estimation~\cite{kang2023efficient}, increasing latency, resource usage, and system complexity.

In this work we advocate treating \textbf{RGB-only inference as a first-class deployment target}, as it represents a low-assumption configuration that avoids auxiliary-modality sensing and preprocessing for practical edge systems~\cite{dvanet}.

Motivated by this, we propose a compact RGB-only action recognition network, which we refer to as X3D-UGT, explicitly designed for embedded execution. Our model builds on an efficient X3D-style backbone augmented with the Temporal Shift Module (TSM) for low-cost temporal modeling~\cite{feichtenhofer2020x3d, lin2019tsm}, and further incorporates a parameter-efficient block structure inspired by mobile architectures (a 3D Universal Inverted Bottleneck with Ghost pointwise convolutions), \textbf{selective temporal adaptation} applied only where it is most effective, and lightweight parameter-free attention for feature refinement. Beyond standard accuracy--efficiency metrics, we evaluate our approach under a unified embedded inference stack and report on-device measurements, including engine size, peak host memory, power consumption, and runtime device memory. Going further, we characterize the frozen model as a perception component for resource-constrained assistive monitoring: we evaluate the trained checkpoints, without any retraining or adaptation, on the datasets' official Medical Conditions action groups and under controlled visual stress. This is a controlled pre-deployment characterization, not a clinical or real-world robot validation.

Our contributions can be summarized as follows.
\begin{itemize}
  \item We present a compact RGB-only action-recognition network that integrates a 3D Universal Inverted Bottleneck, Ghost pointwise convolutions, factorized depthwise operators, temporal shift, selective temporal adaptation, and parameter-free attention.
  \item We demonstrate a favorable accuracy--footprint trade-off on NTU RGB+D 60 and 120 with $0.96$--$0.99$M parameters and provide TensorRT FP16 engine-level measurements on Jetson Orin Nano.
  \item We introduce an inference-only assistive evaluation of frozen checkpoints using the official NTU Medical Conditions groups in the original 60-/120-way decision spaces, including class-balanced recognition and threshold-free clip-level category screening.
  \item We quantify sensitivity to controlled low light, motion blur, partial occlusion, and crop perturbation, showing that, among the four selected Level-2 perturbations, occlusion causes the largest observed degradation for assistive-relevant actions.
\end{itemize}

\section{Related Work}
\label{sec:relatedwork}
\subsection{RGB-only Action Recognition}
RGB-only action recognition has steadily advanced~\cite{tran2015, baradel2018, zhu2019, vyas2020, cheng2022} without relying on depth or skeleton at inference. Self-supervised pretraining targets more transferable spatio-temporal features: ViewCLR~\cite{viewclr} learns representations via contrastive objectives with feature-space transformations for robustness to distribution shift. Supervised contrastive formulations better structure the embedding space; Shah et al.~\cite{shah2023} form semantically consistent positives and separate hard negatives. DVANet~\cite{dvanet} disentangles action-relevant from nuisance factors via a transformer-style decoder, while VT-LVLM-AR~\cite{li2025} adapts a largely frozen LVLM through prompt-tuned, compact temporal event representations. Many such methods, however, are hard to run under tight latency and compute budgets---contrastive learning, disentanglement modules, or LVLM adapters add training or inference overhead---and do not necessarily preserve favorable accuracy--footprint trade-offs on embedded hardware.

\subsection{Lightweight Action Recognition}
Lightweight action recognition targets low latency, limited memory, and restricted compute, where conventional 3D-CNN pipelines are difficult to use. One direction replaces 3D convolutions with 2D backbones plus inexpensive temporal mixing: EPAM-Net~\cite{epam} integrates the Temporal Shift Module (TSM)~\cite{lin2019tsm} into an efficient 2D CNN (X-ShiftNet) and adds a pose-driven spatiotemporal attention block. A second direction exploits richer modalities only at training time: MMCL~\cite{liu2024multi} co-learns with auxiliary instructive signals and discards them at test time, and SV-data2vec~\cite{sv} uses a student--teacher masked-prediction scheme with contextualized latent targets.

Auxiliary modalities enter such pipelines in two ways. MMCL~\cite{liu2024multi} and SV-data2vec~\cite{sv} use them \emph{only during training} and then perform single-modality RGB inference (no inference-time cost, though training still depends on skeleton/pose supervision). Methods that consume an auxiliary modality \emph{at inference}---e.g., a pose stream estimated per clip---instead incur genuine on-device overhead. We target single-stream RGB recognition: the classifier consumes no auxiliary modality at training or inference. As detailed in our setup, the benchmark preprocessing still uses pose-derived person-centric crops following common NTU practice---a fixed input-preparation step whose cost we exclude from on-device measurements, not a modality the network consumes.

\subsection{Efficient Architectural Primitives}
Reusable primitives improve accuracy--efficiency trade-offs with minimal overhead. Squeeze-and-Excitation~\cite{hu2018squeeze} recalibrates channels via global pooling and lightweight gating; GhostNet~\cite{han2020ghostnet} synthesizes most feature maps with cheap operations to cut pointwise-convolution cost; and SimAM~\cite{yang2021simam} derives parameter-free attention weights from an energy formulation. MobileNetV4~\cite{qin2024mobilenetv4} introduces the Universal Inverted Bottleneck (UIB) as a hardware-aware block; TAdaConv~\cite{huang2022tada} calibrates convolution responses by temporal context; and the $(2{+}1)$D factorization~\cite{tran2018closer} separates temporal from spatial processing while preserving receptive-field coverage. We compose a compact backbone from these parameter-efficient blocks and lightweight attention.

\subsection{Edge Deployment, Embedded Inference, and Assistive Monitoring}
Action recognition is increasingly targeted at edge and embedded platforms~\cite{liu2021acdnet, cobparro2024, wang2025rthare}, where inference on the Jetson family is typically served through TensorRT, which fuses operators into a serialized engine~\cite{jeong2022tensorrt}. Deployment behavior depends not only on FLOPs and parameters but also on operator granularity, fusion, memory-access patterns, and kernel-launch overhead~\cite{chen2018tvm, luo2025survey}: large compute-dense operators map well onto Tensor Cores, whereas many small memory-bound operators are limited by dispatch latency and bandwidth. We therefore report deployment-critical metrics---serialized engine size, peak host memory, power, and execution-context memory---alongside accuracy (Sec.~\ref{sec:experiments}).

Vision-based assistive systems span activity and event recognition, pose analysis, and behavioral monitoring~\cite{leo2018assistive}, including activities of daily living in smart homes and assisted-living settings~\cite{debes2016monitoring,das2019toyota,crispim2017behave}; fine-grained ADL recognition must further distinguish visually similar actions with subtle temporal differences~\cite{das2020looking,das2022vpnpp}. Unlike these, our study introduces no complete monitoring pipeline---it is an inference-only reliability characterization of a compact frozen classifier under controlled visual degradation.

  \begin{figure}[t]
    \centering\resizebox{\textwidth}{!}{%
    \includegraphics[scale=1]{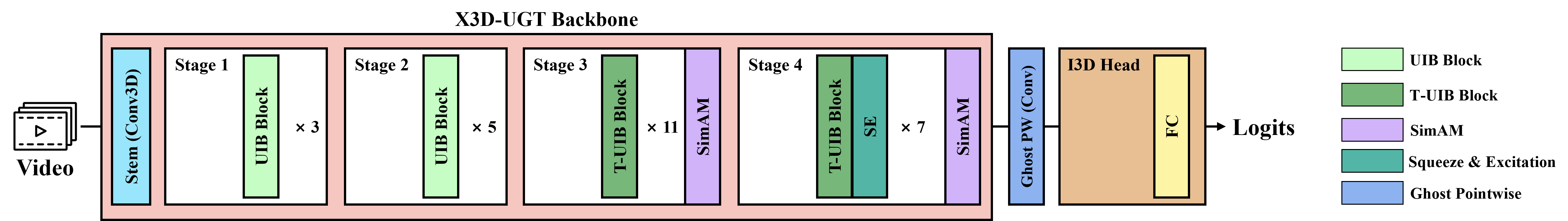}}
    \caption{Overview of the proposed RGB-only network, X3D-UGT: a stem feeds four UIB-based stages. Blocks with selective temporal adaptation (TAda) are denoted T-UIB and occur only in the first two blocks of Stages~3 and~4; parameter-free SimAM follows Stages~3 and~4, and squeeze-and-excitation is used in all Stage~4 blocks. Per-stage block counts are listed in Table~\ref{tab:arch}.}
    \label{fig:fig1}
  \end{figure}

  \begin{figure}[t]
  \resizebox{\columnwidth}{!}{%
    \includegraphics[scale=1]{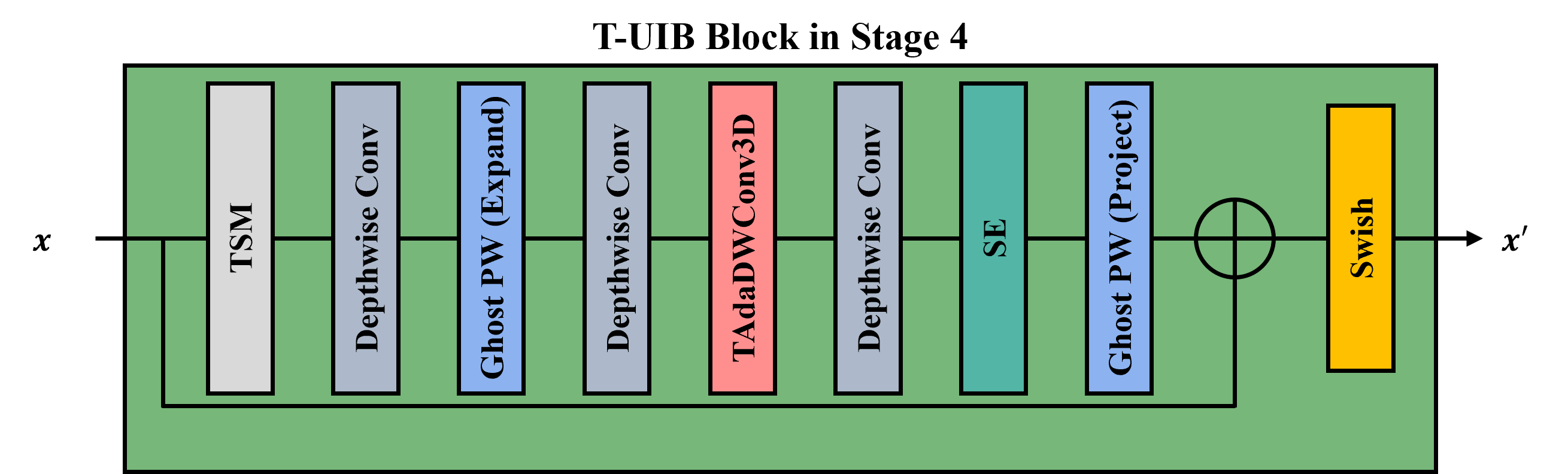}}
    \caption{Stage~4 T-UIB block with explicit kernels. Operator order: temporal shift (TSM); pre-depthwise ($1{\times}5{\times}5$); Ghost pointwise expand; $(2{+}1)$D factorized depthwise (temporal DW/TAda $3{\times}1{\times}1$, then spatial DW $1{\times}3{\times}3$); squeeze-and-excitation; Ghost pointwise project; residual add.}
    \label{fig:fig2}
  \end{figure}

\section{Method}
\label{sec:method}

\subsection{Overall Architecture}
Our network is based on the X3D-style hierarchical design~\cite{feichtenhofer2020x3d} combined with the Temporal Shift Module (TSM)~\cite{lin2019tsm} for low-cost temporal modeling. As illustrated in Fig.~\ref{fig:fig1}, the architecture consists of a stem layer, four stages, and a classification head. The stem applies a $(1 \times 3 \times 3)$ convolution with stride $(1 \times 2 \times 2)$ to reduce spatial resolution while preserving temporal information; no temporal downsampling is performed at the stem. Subsequent stages perform progressive spatial downsampling and channel expansion, followed by a $1 \times 1 \times 1$ projection (conv5), global average pooling, and classification. Table~\ref{tab:arch} summarizes the stage-wise configuration, including the number of blocks, channel widths, spatial strides, and where each primitive is enabled.

To balance efficiency and recognition performance, we integrate four complementary design choices: (1) a \textit{parameter-efficient block structure} adapted from mobile architectures; (2) \textit{selective temporal adaptation} applied only where it is most effective; (3) \textit{lightweight, parameter-free attention} for improved feature discrimination; and (4) a \textit{lightweight training objective} that enhances convergence without increasing inference cost. The architecture instantiates a single block type, a 3D UIB, across all four stages; what differs across stages is which optional primitives (temporal adaptation, squeeze-and-excitation, attention) are switched on. This keeps the architectural definition uniform across stages while still allocating capacity where it is most useful.

\begin{table}[t]
\centering
\caption{Stage-wise configuration of the proposed network. ``blk 0,1'' denotes TAda enabled only in the first two blocks of the stage; SimAM is applied at the stage output and SE in all Stage-4 blocks. The head applies adaptive average pooling, dropout ($0.5$), and a linear classifier on the $432$-dim feature.}
\label{tab:arch}
\small
\setlength{\tabcolsep}{4pt}
\begin{tabular}{lccccccc}
\toprule
Stage & \#blk & ch (in$\to$out) & stride & TAda & SimAM & SE & Ghost \\
\midrule
Stem  & --  & $3\to24$    & $(1,2,2)$ & --        & --        & --        & --        \\
S1    & 3   & $24\to24$   & $(1,1,1)$ & \ding{55} & \ding{55} & \ding{55} & \checkmark \\
S2    & 5   & $24\to48$   & $(1,2,2)$ & \ding{55} & \ding{55} & \ding{55} & \checkmark \\
S3    & 11  & $48\to96$   & $(1,2,2)$ & blk 0,1   & after S3  & \ding{55} & \checkmark \\
S4    & 7   & $96\to192$  & $(1,2,2)$ & blk 0,1   & after S4  & all blk   & \checkmark \\
conv5 & --  & $192\to432$ & --        & --        & --        & --        & \checkmark \\
\bottomrule
\end{tabular}
\end{table}

The per-stage output tensor shapes for a $3\times16\times224\times224$ RGB clip---channels expanding $24\to48\to96\to192$ before a $432$-dimensional projection, with spatial resolution halved at the stem and at the first block of Stages~2--4 and temporal length preserved ($T{=}16$)---are tabulated in Appendix~\ref{app:shapes}.

\subsection{Efficient Building Blocks}
\textbf{Universal Inverted Bottleneck (UIB) for 3D features.}
The inverted bottleneck follows an expand--depthwise--project pattern, where channels are first expanded, then processed by a depthwise operator, and finally projected back to a compact representation. We adapt the Universal Inverted Bottleneck (UIB) design from MobileNetV4~\cite{qin2024mobilenetv4} to 3D video features by introducing an optional pre-depthwise spatial convolution that operates on the input channels prior to expansion. This pre-depthwise layer uses a $(1 \times 5 \times 5)$ spatial-only kernel to refine spatial features at low cost, and performs spatial downsampling when required, allowing subsequent expanded features to be processed at reduced resolution. The expanded (mid) channel widths follow $\{56, 112, 216, 432\}$ for Stages~1--4, and an effective mid width of $\mathrm{make\_divisible}(\mathrm{mid}\times 0.75, 8)$ is used inside the depthwise stage to further trim cost.

For the mid-depthwise operation, we adopt the $(2{+}1)$D factorization strategy~\cite{tran2018closer}, decomposing depthwise 3D convolution into a $(3 \times 1 \times 1)$ temporal convolution followed by a $(1 \times 3 \times 3)$ spatial convolution; spatial striding is applied within the spatial component. This factorization explicitly separates temporal mixing from spatial processing, reducing computational cost while preserving receptive-field coverage. As illustrated in Fig.~\ref{fig:fig2}, each block first applies temporal shift, then the inverted-bottleneck transformation, followed by projection and residual addition. The full forward order of a block is $\textsf{TSM} \to \textsf{pre-dw} \to \textsf{pw-expand (Ghost)} \to \textsf{mid-dw } (2{+}1)\textsf{D} \to [\textsf{SE: Stage 4}] \to \textsf{pw-project (Ghost)} \to +\textsf{residual} \to \textsf{Swish}$, where the residual path uses a Ghost downsampling convolution when the spatial resolution or channel count changes. Squeeze-and-excitation~\cite{hu2018squeeze} is applied only in Stage~4 to maintain a favorable accuracy--footprint balance.

\noindent\textbf{Temporal Shift.}
Following TSM~\cite{lin2019tsm}, the first operation of every block shifts a fraction of channels along the temporal axis: with $n_{\text{div}}=8$, one eighth of the channels are shifted forward and one eighth backward (one quarter in total), while the remainder are left untouched. Temporal shift introduces no parameters and negligible arithmetic, yet enables information exchange across neighboring frames before the spatial operators act, which is why we place it at the very beginning of each block.

\noindent\textbf{Ghost pointwise convolution.}
Pointwise $(1 \times 1 \times 1)$ convolutions dominate the parameter count in inverted-bottleneck architectures due to channel expansion. To mitigate this overhead, we extend the Ghost module~\cite{han2020ghostnet} to 3D by replacing standard pointwise convolutions with a two-branch structure. A primary $1 \times 1 \times 1$ convolution generates a subset of output channels (ratio $2$, i.e., half), while the remaining channels are produced by a cheap depthwise convolution with a $(1 \times 3 \times 3)$ kernel applied to the primary output; striding, when needed, is applied only in the primary branch. As shown in Fig.~\ref{fig:fig2}, this Ghost-based replacement is used for the expansion and projection layers within each block, for residual downsampling, and for the final feature projection (conv5). Because pointwise layers are where most parameters concentrate, this substitution is a major contributor to the model's sub-1M parameter budget.

\subsection{Selective Temporal Adaptation}
Temporal Adaptive Convolution (TAdaConv)~\cite{huang2022tada} calibrates convolution responses based on temporal context and has shown strong performance in video understanding. However, applying temporal adaptation uniformly across all layers incurs noticeable overhead, as each adapted layer introduces additional gating whose complexity grows with channel width. We observe that temporal adaptation is not equally beneficial throughout the network: early layers primarily capture low-level spatial cues such as edges and textures that vary little across frames, whereas deeper layers encode semantic representations where temporal dynamics are more critical for distinguishing actions.

Motivated by this observation, we restrict temporal adaptation to a small number of blocks in the deeper stages. Specifically, TAda is enabled only in the first $K$ blocks of Stage~3 and Stage~4 (with $K=2$ by default), as highlighted in Fig.~\ref{fig:fig1}; that is, the adapted blocks are $(3,0), (3,1), (4,0), (4,1)$. Following TAdaConv, rather than generating fully dynamic convolution kernels, we adopt a simplified variant that performs channel-wise residual scaling after the temporal depthwise convolution: the $(3 \times 1 \times 1)$ temporal depthwise component is replaced by a temporally adaptive variant. The gating signal is derived from temporally aggregated context using a lightweight bottleneck (reduction $16$): global spatial pooling, a $1{\times}1$ channel-reduction convolution, Swish, a depthwise temporal convolution with kernel $3$, a $1{\times}1$ channel-expansion convolution, and a sigmoid. The modulation is applied as $y \leftarrow y \cdot \big(1 + \alpha\,(g - 1)\big)$, where $g$ is the gate and the learnable blending coefficient $\alpha$ is initialized to zero, so that the block begins as an identity mapping and the network gradually incorporates temporal adaptation during training. This identity-initialized, deep-stage-only design keeps the added cost small and, as our study in Sec.~\ref{sec:ablation} shows, matches or exceeds the all-stage variant while using fewer parameters.

\subsection{Parameter-Free Attention and Training Objective}
\textbf{Parameter-free attention.}
To improve feature discrimination without adding learnable parameters, we incorporate SimAM~\cite{yang2021simam} as a lightweight attention mechanism. We apply SimAM to the high-level outputs of Stage~3 and Stage~4, adding negligible computational overhead due to its parameter-free, energy-based, element-wise formulation. We note that this placement (after both Stage~3 and Stage~4) corresponds to the public training configuration used for our reported results, rather than the code default, and we make this explicit for reproducibility.

\noindent\textbf{Training objective.}
We train our network using the Poly-1 loss~\cite{leng2022polyloss}, which augments standard cross-entropy with a polynomial term:
\begin{equation}
\mathcal{L}_{\text{Poly-1}} = \mathcal{L}_{\text{CE}}(p, y) + \epsilon \cdot (1 - p_y),
\end{equation}
where $\epsilon$ is a hyperparameter and $p_y$ denotes the predicted probability of the ground-truth class. This term strengthens gradients for samples whose ground-truth probability is low, improving convergence while leaving inference unchanged. As we report in Sec.~\ref{sec:ablation}, the resulting improvement is modest, and we therefore treat Poly-1 as an inference-free training choice rather than a primary source of the overall gain.

\begin{table}[t]
\centering
\caption{Comparison with state-of-the-art \textbf{RGB-only} action recognition methods on NTU RGB+D 60 and 120 (RGB modality only). \textit{Italic} values for EPAM-Net are its single-modality RGB-stream (RGB-X-ShiftNet) results from the original paper. ``Ours'' parameter counts are $0.96$M (NTU60) and $0.99$M (NTU120).}
\label{table1}
\resizebox{\linewidth}{!}{%
\begin{tabular}{|l||cc|cc|c|c|}
\hline
\multirow{2}{*}{Model}            & \multicolumn{2}{c|}{NTU 60}         & \multicolumn{2}{c|}{NTU 120}       & \multirow{2}{*}{GFLOPs} & \multirow{2}{*}{Params (M)} \\ \cline{2-5}
                                  & \multicolumn{1}{c|}{X-Sub} & X-View & \multicolumn{1}{c|}{X-Sub}  & X-Set  &                         &                            \\ \hline\hline
(Tran et al. 2015) (ICCV '15)~\cite{tran2015}     & \multicolumn{1}{c|}{63.5}  & 70.3   & \multicolumn{1}{c|}{-}     & -     & 38.5                    & 78.4                       \\ \hline
(Baradel et al. 2018) (CVPR '18)~\cite{baradel2018} & \multicolumn{1}{c|}{86.6}  & 93.2   & \multicolumn{1}{c|}{-}     & -     & -                       & -                          \\ \hline
(Zhu et al. 2019) (SPL '19)~\cite{zhu2019}     & \multicolumn{1}{c|}{93.2}  & 97.7   & \multicolumn{1}{c|}{-}     & -     & 51.7                    & 33.0                       \\ \hline
(Vyas et al. 2020) (ECCV '20)~\cite{vyas2020}     & \multicolumn{1}{c|}{88.9}  & 86.3   & \multicolumn{1}{c|}{-}     & -     & -                       & -                          \\ \hline
(Cheng et al. 2022)~\cite{cheng2022}               & \multicolumn{1}{c|}{91.9}  & 95.4   & \multicolumn{1}{c|}{-}     & -     & -                       & -                          \\ \hline
ViewCLR (WACV '23)~\cite{viewclr}                & \multicolumn{1}{c|}{89.7}  & 94.1   & \multicolumn{1}{c|}{84.5}  & 86.2  & -                       & -                          \\ \hline
(Shah et al. 2023) (WACV '23)~\cite{shah2023}     & \multicolumn{1}{c|}{91.4}  & 98.0   & \multicolumn{1}{c|}{85.6}  & 87.5  & -                       & -                          \\ \hline
DVANet (AAAI '24)~\cite{dvanet}                 & \multicolumn{1}{c|}{93.4}  & 98.2   & \multicolumn{1}{c|}{90.4}  & 91.6  & 25.5                    & 40.5                       \\ \hline
VT-LVLM-AR (Li et al. 2025)~\cite{li2025}                  & \multicolumn{1}{c|}{94.1}  & 96.8   & \multicolumn{1}{c|}{87.0}  & 88.5  & -                       & -                          \\ \hline
EPAM-Net (RGB stream)~\cite{epam}     & \multicolumn{1}{c|}{\textit{94.83}}  & \textit{98.04}  & \multicolumn{1}{c|}{\textit{90.13}} & \textit{91.58} & 4.5                     & 2.0                        \\ \hline\hline
Ours                              & \multicolumn{1}{c|}{\textbf{95.10}} & \textbf{98.31}  & \multicolumn{1}{c|}{\textbf{90.88}} & \textbf{92.66} & 9.15                       & \textbf{0.96--0.99}                       \\ \hline
\end{tabular}}
\end{table}

\begin{table}[t]
\centering
\caption{Stage-wise placement of selective temporal adaptation (TAda) on NTU60; in every configuration TAda is enabled only in the \emph{first two blocks} of each listed stage. ``Deep stages (S3--S4)'' is our final model. Single-run configuration-study results.}
\label{tab:tada_ablation}
\small
\begin{tabular}{lccc}
\toprule
TAda placement & Params (K) & NTU60 X-Sub & NTU60 X-View \\
\midrule
All stages (S1--S4)   & 965 & 94.97 & 98.29 \\
Early stages (S1--S2) & 930 & 95.08 & 98.23 \\
S3 only               & 935 & 95.06 & 98.16 \\
S4 only               & 955 & 94.87 & 98.19 \\
Deep stages (S3--S4)  & 962 & \textbf{95.21} & \textbf{98.31} \\
\bottomrule
\end{tabular}
\end{table}

\subsection{Implementation Details}
\label{sec:impl}
All experiments are implemented in PyTorch and trained with Distributed Data Parallel (DDP, NCCL backend) on a single node. Optimization uses SGD with an initial learning rate of $0.05$, momentum $0.9$, and weight decay $3\times10^{-4}$, under a CosineAnnealingLR schedule that is stepped \emph{per iteration} with $T_{\max} = \lvert \text{train loader} \rvert \times \text{epochs}$. We apply gradient clipping with a maximum $\ell_2$ norm of $40$ at every iteration. We train for $200$ epochs. We use Poly-1 cross-entropy with $\epsilon=1.0$ for all final models, except the Poly-1 ablation, which uses plain cross-entropy.

Each input clip is uniformly sampled to $16$ frames (via \texttt{np.linspace}) and resized to $224\times224$. Training-time augmentation consists of $1/255$ scaling, a clip-consistent horizontal flip with probability $0.5$, and a per-frame brightness jitter in $[0.8, 1.2]$; we do not use mean/std normalization, mixup, cutmix, random spatial crops, or rotations. During preprocessing we apply a person-centric crop. Person-centric normalization is standard in NTU pipelines---skeleton-based methods such as Hyperformer~\cite{zhou2022hypergraph} re-center each sequence about the body (spine) joint---and for the RGB modality we follow EPAM-Net~\cite{epam}: 2D pose keypoints are used to compute a single per-clip ``global'' bounding box that encloses the subject(s) across all frames of the clip, and every frame is then cropped to this box and resized to $224\times224$. We state the scope of our RGB-only claim precisely: the recognition backbone consumes only RGB pixels and uses no auxiliary pose or skeleton stream at training or inference, but the benchmark preprocessing is \emph{not} pose-free---the pose keypoints serve only to localize the crop region. This crop is a one-time, offline input-preparation step, and the cost of obtaining it is excluded from our on-device latency and energy measurements. The per-GPU batch size is $8$; the final headline models use $2$ GPUs (global batch $16$), and the Poly-1 ablation run uses $6$ GPUs (global batch $48$). All reported numbers correspond to single training runs; we do not fix random seeds and therefore do not report variance across seeds.

\begin{table}[t]
\centering
\caption{Deployment-oriented on-device characterization on \textbf{Jetson Orin Nano} (MAXN SUPER mode) using a unified TensorRT FP16 inference pipeline. S denotes the skeleton modality. \textbf{Acc$^{\ast}$} is offline NTU top-1 accuracy, reported for reference only and not part of the on-device timing loop; throughput, engine size, memory, and power are measured on-device. Underlined/bold entries mark the best two values per column.}
\label{table3}
\resizebox{\textwidth}{!}{%
\begin{tabular}{|l|l|c|c||c|c|c|c|c|}
\hline
Model      & Modality & Params $\downarrow$    & Acc$^{\ast}$ $\uparrow$          & Thr.\ (clips/s) $\uparrow$           & Eng.\ (MiB) $\downarrow$ & RAM (MB) $\downarrow$ & Power (W) $\downarrow$     & \begin{tabular}[c]{@{}c@{}}TRT Ctx \\ Mem (MiB) $\downarrow$\end{tabular} \\ \hline\hline
EPAM-Net~\cite{epam}       & RGB      & {\ul 2.06}    & 96.5$^{\ast}$         & 23.350          & {\ul 19.92}   & \textbf{3,141}  & \textbf{12.824} & \textbf{80.21}                                                      \\ \hline
DVANet~\cite{dvanet}      & RGB      & 40.5         & 98.2$^{\ast}$          & {\ul 29.192}    & 69.77         & 3,292           & 19.371          & 73.50                                                                 \\ \hline
SVdata2vec~\cite{sv} & RGB+S    & 18.4         & \textbf{99.3$^{\ast}$} & \textbf{51.259} & 37.04         & 3,248           & 17.621          & 137.84                                                                \\ \hline
Ours       & RGB      & \textbf{0.96} & {\ul 98.3$^{\ast}$}   & 10.305          & \textbf{5.30} & {\ul 3,153}     & {\ul 16.947}    & {\ul 89.89}                                                           \\ \hline
\end{tabular}
}
\end{table}

\section{Experiments}
\label{sec:experiments}
\subsection{Datasets and Protocols}
We evaluate on NTU RGB+D 60~\cite{shahroudy2016ntu} and NTU RGB+D 120~\cite{liu2020ntu120} under their standard splits. For NTU60 we use Cross-Subject (X-Sub) and Cross-View (X-View); for NTU120 we use Cross-Subject (X-Sub) and Cross-Setup (X-Set). The corresponding test-set sizes in our setup are $16{,}487$ (NTU60 X-Sub), $18{,}932$ (NTU60 X-View), $50{,}919$ (NTU120 X-Sub), and $59{,}477$ (NTU120 X-Set). Because the protocols differ in subject, view, and setup partitions, accuracies are comparable only within a protocol; we therefore avoid cross-protocol numerical comparisons. The NTU60 models contain $962{,}544$ parameters ($0.96$M) and the NTU120 models contain $988{,}524$ ($0.99$M); the difference of $25{,}980$ parameters arises solely from the larger classification head. For assistive-oriented analysis, we use the Medical Conditions categories defined in the original NTU RGB+D taxonomies~\cite{shahroudy2016ntu,liu2020ntu120}. Med9 contains actions A41--A49 in NTU60, while Med12 contains A41--A49 and A103--A105 in NTU120. All metrics retain the original 60- or 120-class prediction space; we never re-normalize logits over the medical subset.

\subsection{Comparisons to State-of-the-Art}
We compare our method against representative \textbf{RGB-only} action recognition methods on NTU RGB+D 60 and 120, as reported in Table~\ref{table1}. All methods are evaluated under RGB-only inference. For EPAM-Net, we cite the RGB-stream results reported in the original paper.

Our method achieves strong performance across all protocols, reaching $\mathbf{95.10\%}$ and $\mathbf{98.31\%}$ on NTU60 (X-Sub/X-View) and $\mathbf{90.88\%}$ and $\mathbf{92.66\%}$ on NTU120 (X-Sub/X-Set). Compared with the single-modality RGB-stream results reported by EPAM-Net ($94.83/98.04$ on NTU60 and $90.13/91.58$ on NTU120), our model is higher on every protocol---by $+0.27$ and $+0.27$ points on NTU60 (X-Sub/X-View) and by $+0.75$ and $+1.08$ points on NTU120 (X-Sub/X-Set)---while using only $0.96$--$0.99$M parameters, less than half of the EPAM-Net RGB backbone. Although our model uses fewer parameters, its nominal computation of $9.15$ GFLOPs is higher than EPAM-Net's $4.5$ GFLOPs; we therefore claim parameter and storage compactness rather than FLOP or latency superiority. We note that the full multimodal EPAM-Net (RGB+pose) attains higher accuracy ($96.1/99.0/92.4/94.3$), but it relies on an auxiliary pose stream as a network input at inference, whereas our recognition backbone consumes no such stream. The combination of higher accuracy than the comparable RGB-only baseline and a smaller parameter budget is the central message of Table~\ref{table1}: under the single-stream RGB-only setting, our method offers a favorable accuracy--footprint trade-off that aligns well with embedded use.

\subsection{Assistive-Oriented Inference-Only Evaluation}
\label{sec:assistive}
\noindent\textbf{Protocol and metrics.}
We evaluate the preserved checkpoints with frozen weights and no retraining, fine-tuning, or test-time adaptation. Medical-condition recognition is evaluated within the original label space using medical-subset top-1, macro recall, macro F1, and worst-class recall. We additionally define a clip-level medical-category score as the sum of softmax probabilities assigned to the medical classes,
\begin{equation}
s_{\mathrm{med}}(x)=\sum_{c\in\mathcal{M}}p(y=c\mid x),
\end{equation}
where $\mathcal{M}$ is the official Medical Conditions group. Because the repository contains no independent validation split, we report threshold-free AUROC and AUPRC for this clip-level category screening and do not select operating thresholds or report FNR at a fixed FPR. This is clip-level category screening, not continuous event detection.

\noindent\textbf{Clean assistive-oriented results.}
The compact model retains strong recognition on the assistive-oriented groups while operating in the full label space (Table~\ref{tab:clean_assistive}). On NTU60, medical macro recall is $93.19\%$ on X-Sub and $97.93\%$ on X-View, with worst-class recalls of $84.73\%$ and $94.62\%$. On NTU120, medical macro recall is $91.47\%$ on X-Sub and $91.48\%$ on X-Set. The summed medical-class probability produces AUROC values between $0.9972$ and $0.9998$, showing that the category is highly separable on clean, trimmed clips. This threshold-free result should not be interpreted as continuous-event detection performance.

\begin{table}[t]
\centering
\caption{Assistive-oriented clean evaluation of the frozen compact RGB-only model on the NTU Medical Conditions groups, retaining the original 60-/120-way decision spaces; no retraining or adaptation is used. Med.\ Macro R: medical macro recall; Worst R: worst-class recall. Medical subsets contain 2{,}481/2{,}844 clips for NTU60 X-Sub/X-View and 4{,}207/6{,}007 for NTU120 X-Sub/X-Set.}
\label{tab:clean_assistive}
\small
\begin{tabular}{llccccc}
\toprule
Dataset & Split & Overall & Med.\ Macro R & Worst R & AUROC & AUPRC \\
\midrule
NTU60  & X-Sub  & 95.10 & 93.19 & 84.73 & 0.9974 & 0.9884 \\
NTU60  & X-View & 98.31 & 97.93 & 94.62 & 0.9998 & 0.9991 \\
NTU120 & X-Sub  & 90.88 & 91.47 & 83.48 & 0.9972 & 0.9790 \\
NTU120 & X-Set  & 92.66 & 91.48 & 82.72 & 0.9985 & 0.9894 \\
\bottomrule
\end{tabular}
\end{table}

\noindent\textbf{Controlled visual stress and failure analysis.}
We then subject the same frozen checkpoint to controlled, inference-only visual stress. Corruptions are applied to the post-crop, $16$-frame, $224\times224$, $[0,1]$ RGB tensor immediately before inference, with one parameter set applied consistently to all $16$ frames of a clip. The Level-2 settings are: low light (gamma $2.0$); motion blur ($9$-pixel line kernel at a deterministic random angle); occlusion (a black rectangle covering $20\%$ of the image area at a deterministic location and aspect ratio); and a spatial crop-perturbation proxy (translation and scale each within $\pm10\%$, bilinear interpolation, zero padding). Motion blur, occlusion, and crop perturbation use three deterministic seeds; low light is deterministic. No adaptation is performed, and this is synthetic controlled stress rather than real-world robot validation.

At Level 2 on NTU60 X-Sub (Table~\ref{tab:visual_stress_l2}), medical macro recall falls only modestly under low light ($1.77$ points) and motion blur ($2.23$ points), and by $4.87$ points under crop perturbation. Occlusion is substantially more damaging, reducing recall by $19.05$ points to $74.14\%$. The overall top-1 results show the same ordering. The X-View results (Appendix~\ref{app:stress_extra}) are consistent, with occlusion again producing the largest medical-recall loss. Among the four tested perturbations, these findings indicate that visibility loss from occlusion---rather than moderate photometric or blur stress---produces the largest observed degradation of the frozen model in this controlled setting. Real-world home activity understanding is further affected by occlusion, non-optimal viewpoints, and imperfect illumination~\cite{rai2021homage,das2019toyota,chen2025opph}, which motivates characterizing these factors explicitly.

\begin{table}[t]
\centering
\caption{Inference-only Level-2 visual stress test on NTU60 X-Sub using the frozen checkpoint. Motion blur, occlusion, and crop perturbation are averaged over three deterministic seeds (mean $\pm$ standard deviation). All values are percentages; drops are absolute percentage points.}
\label{tab:visual_stress_l2}
\resizebox{\linewidth}{!}{%
\begin{tabular}{lccccccc}
\toprule
Metric & Clean & Low light & Motion blur & Occlusion & Crop perturb. & Corr.\ mean & Mean drop \\
\midrule
Overall Top-1   & 95.10 & 93.73 & $92.11\pm0.10$ & $76.97\pm0.14$ & $91.27\pm0.11$ & 88.52 & 6.58 \\
Medical Macro R & 93.19 & 91.41 & $90.96\pm0.32$ & $74.14\pm0.73$ & $88.31\pm0.31$ & 86.21 & 6.98 \\
\bottomrule
\end{tabular}}
\end{table}

A class-level breakdown (Appendix~\ref{app:classlevel}) reveals heterogeneous failure patterns: falling down remains highly recognizable even under occlusion, whereas sneeze/cough and nausea/vomiting are the weakest classes.

\subsection{Ablation Study}
\label{sec:ablation}
We study the placement of selective temporal adaptation while keeping the training protocol fixed. Table~\ref{tab:tada_ablation} reports NTU60 X-Sub and X-View accuracy as we vary which stages enable TAda (in all cases only in the first two blocks of each stage). The absolute differences are small (within roughly $0.1$--$0.3$ percentage points). The selected deep-stage placement (S3--S4) produced the highest observed single-run accuracy on both protocols, at a compact $962$K parameters. The $95.21\%$ shown here is a single configuration-study run; the checkpoint deployed throughout the rest of the paper reproduces $95.10\%$ overall top-1 on NTU60 X-Sub under full test-set re-inference, and this $95.10\%$ is the value used in all headline and assistive results. Given the small differences and the absence of repeated runs, we interpret this as configuration sensitivity that guided our final design choice rather than statistical evidence that deep-stage placement is inherently superior.

We also examined Poly-1 loss, but the observed difference was small (about $0.1$ percentage point on NTU60 X-Sub) and the runs differed in effective batch size. We therefore treat Poly-1 as a secondary, inference-free training choice and do not attribute the final performance gain to it.

\subsection{Edge Deployability Analysis}

The full protocol used to obtain the Jetson measurements in Table~\ref{table3}---hardware/software environment, TensorRT engine construction, and throughput, memory, and power measurement---is detailed in Appendix~\ref{app:protocol} and summarized in Table~\ref{tab:protocol}.

Table~\ref{table3} reports the resulting TensorRT engine-level on-device characterization; the \textbf{Acc$^{\ast}$} column is taken from offline NTU RGB+D evaluation and is included for reference only, as it is not measured inside the timing loop. The EPAM-Net entry ($96.5\%$) is our own re-implementation of EPAM-Net evaluated through the same TensorRT FP16 engine on the Jetson, and is therefore distinct from the published EPAM-Net RGB-stream result in Table~\ref{table1}; the DVANet and SVdata2vec accuracies are quoted from their original papers. Our model achieves the smallest parameter count ($0.96$M) and most compact engine ($5.30$\,MiB)---about $3.7\times$ smaller than EPAM-Net and $13\times$ smaller than DVANet---while retaining strong offline accuracy ($98.3\%$ on NTU60 X-View), competitive with DVANet ($98.2\%$) at only $2.4\%$ of its parameters. Its peak system RAM ($3{,}153$MB) and mean \texttt{VDD\_IN} power ($16.947$W) fall within the observed range, and its execution-context memory ($89.89$MiB) stays well below SVdata2vec, indicating a manageable GPU-memory footprint.

These are engine-level measurements (host-to-device transfer, TensorRT execution, device-to-host transfer), not end-to-end: video decoding, sampling, cropping, resizing, normalization, and postprocessing are excluded (Appendix~\ref{app:protocol}).

We claim no runtime superiority: at $10.305$ clips/s our model is the slowest of the four, so static compactness does not translate into higher TensorRT throughput on this embedded GPU. We therefore position it as a parameter- and storage-efficient RGB-only network for storage-constrained or lower-rate monitoring rather than a high-throughput one.

\subsection{Limitations}
Although our model achieves the smallest parameter count ($0.96$M) and the most compact TensorRT engine ($5.30$\,MiB), its inference throughput on Jetson Orin Nano is lower than that of the baselines. We did not perform operator-level profiling, so we do not claim a specific bottleneck operator; a plausible, structure-consistent explanation---memory-bound shift, Ghost, factorized-depthwise, and TAda operators producing a fragmented graph of many small kernels with limited fusion---is discussed in Appendix~\ref{app:bottleneck}, and verifying it through operator-level profiling is left to future work.

The assistive analysis is diagnostic, not a robustness claim: it uses frozen checkpoints and synthetic corruptions on trimmed NTU clips, crop perturbation is only a spatial proxy, and no detector or end-to-end pipeline is evaluated. With no independent validation split, we report threshold-free screening metrics and select no alert operating point. As the repository has no preserved RGB baseline checkpoint for matched corruption testing, the stress results characterize our model but do not establish a robustness advantage over other architectures. Finally, the Medical Conditions subset is an action-recognition taxonomy, not clinical, continuous-detection, or safety-critical validation; unlike untrimmed ADL benchmarks~\cite{dai2023tsu}, our pre-segmented clips do not measure event onset, duration, alert latency, or false alarms.

\section{Conclusion}
We presented X3D-UGT, a compact RGB-only action-recognition network for resource-constrained assistive monitoring. The model preserves competitive NTU60 and NTU120 accuracy with a sub-1M parameter budget and a $5.30$\,MiB TensorRT engine. Frozen-checkpoint evaluation on the official Medical Conditions groups shows strong clean recognition and category separability, while controlled visual stress identifies occlusion as the largest source of degradation among the tested perturbations. These results support the model's use as a compact, lower-rate perception component and also delineate the reliability limitations that must be addressed before real-world assistive deployment.

\putbib[refs]
\end{bibunit}

\clearpage
\appendix
\begin{bibunit}[splncs04]

\section{Per-Stage Tensor Shapes}
\label{app:shapes}
Table~\ref{tab:shapes} makes the spatial--temporal resolution and channel schedule explicit for a $3\times16\times224\times224$ RGB clip. The temporal length is preserved throughout ($T{=}16$; no temporal downsampling), spatial resolution is halved at the stem and at the first block of Stages~2--4, and channels expand $24\to48\to96\to192$ before a $432$-dimensional projection feeds the classifier.

\begin{table}[t]
\centering
\caption{Per-stage output tensor shapes for a $3\times16\times224\times224$ input clip.}
\label{tab:shapes}
\small
\begin{tabular}{lc}
\toprule
Stage & Output tensor ($C\times T\times H\times W$) \\
\midrule
Input   & $3\times16\times224\times224$  \\
Stem    & $24\times16\times112\times112$ \\
Stage 1 & $24\times16\times112\times112$ \\
Stage 2 & $48\times16\times56\times56$   \\
Stage 3 & $96\times16\times28\times28$   \\
Stage 4 & $192\times16\times14\times14$  \\
conv5   & $432\times16\times14\times14$  \\
Head    & $432 \to \#\mathrm{classes}$   \\
\bottomrule
\end{tabular}
\end{table}

\section{Class-Level Failure Analysis}
\label{app:classlevel}
Class-level analysis further shows heterogeneous failure patterns. Falling down---a canonical target for vision-based assistive monitoring~\cite{nunezmarcos2017fall}---remains highly recognizable in this protocol ($100.0\%$ clean recall and $93.7\%$ under Level-2 occlusion), whereas sneeze/cough and nausea/vomiting have the lowest clean recalls, both near $84.7\%$. Headache, sneeze/cough, and neck pain exhibit the largest average degradation across the tested corruptions. Full per-class clean and Level-2 recall values are reported in Table~\ref{tab:per_class}.

\begin{table}[t]
\centering
\caption{Per-class medical recall on NTU60 X-Sub (frozen checkpoint, inference-only). Clean and Level-2 corruption recall (\%) for each Medical Conditions class, with mean corrupted recall and absolute drop (percentage points). Motion blur, occlusion, and crop perturbation are averaged over three deterministic seeds.}
\label{tab:per_class}
\resizebox{\linewidth}{!}{%
\begin{tabular}{lrrrrrrr}
\toprule
Class & Clean & Low light & Motion blur & Occlusion & Crop perturb. & Corr.\ mean & Drop \\
\midrule
Sneeze/cough     & 84.78 & 80.43 & 82.61 & 60.63 & 75.97 & 74.91 & 9.87 \\
Staggering       & 98.19 & 96.74 & 94.44 & 77.42 & 97.46 & 91.52 & 6.67 \\
Falling down     & 100.00 & 100.00 & 99.76 & 93.70 & 100.00 & 98.36 & 1.64 \\
Headache         & 92.39 & 87.68 & 89.61 & 71.14 & 81.64 & 82.52 & 9.87 \\
Chest pain       & 95.29 & 94.57 & 94.32 & 65.34 & 91.30 & 86.38 & 8.91 \\
Back pain        & 98.91 & 99.28 & 96.62 & 90.70 & 96.74 & 95.83 & 3.08 \\
Neck pain        & 92.39 & 89.86 & 89.86 & 69.69 & 82.85 & 83.06 & 9.33 \\
Nausea/vomiting  & 84.73 & 80.73 & 82.18 & 63.52 & 82.06 & 77.12 & 7.61 \\
Fan self         & 92.00 & 93.45 & 89.21 & 75.15 & 86.79 & 86.15 & 5.85 \\
\bottomrule
\end{tabular}}
\end{table}

\section{Additional Visual-Stress Results}
\label{app:stress_extra}
Table~\ref{tab:xview_stress} repeats the Level-2 visual-stress protocol of Table~\ref{tab:visual_stress_l2} on the NTU60 X-View split, and Table~\ref{tab:all_severities} sweeps corruption severity across Levels~1--3 on NTU60 X-Sub. In both, occlusion produces by far the largest degradation, consistent with the X-Sub Level-2 results in the main paper.

\begin{table}[t]
\centering
\caption{Inference-only Level-2 visual stress test on NTU60 \textbf{X-View} using the frozen checkpoint (no retraining or adaptation). Motion blur, occlusion, and crop perturbation are averaged over three deterministic seeds (mean $\pm$ std). Crop perturbation is a spatial crop-perturbation proxy. All values are percentages; drops are absolute percentage points.}
\label{tab:xview_stress}
\resizebox{\linewidth}{!}{%
\begin{tabular}{lrrrrrrr}
\toprule
Metric & Clean & Low light & Motion blur & Occlusion & Crop perturb. & Corr.\ mean & Mean drop \\
\midrule
Overall Top-1   & 98.31 & 97.19 & $94.80\pm0.03$ & $81.06\pm0.25$ & $94.23\pm0.12$ & 91.82 & 6.49 \\
Medical Macro R & 97.93 & 96.66 & $95.73\pm0.23$ & $76.90\pm0.52$ & $92.50\pm0.09$ & 90.45 & 7.48 \\
\bottomrule
\end{tabular}}
\end{table}

\begin{table}[t]
\centering
\caption{All-severity inference-only visual stress on NTU60 X-Sub (frozen checkpoint). Corruption severity is swept across Levels~1--3; motion blur, occlusion, and crop perturbation are averaged over three deterministic seeds (mean $\pm$ std). All values are percentages; mean drop is the absolute percentage-point drop averaged over the four perturbations.}
\label{tab:all_severities}
\resizebox{\linewidth}{!}{%
\begin{tabular}{llrrrrrr}
\toprule
Metric & Sev. & Clean & Low light & Motion blur & Occlusion & Crop perturb. & Mean drop \\
\midrule
Overall Top-1         & L1 & 95.10 & 94.66 & $94.31\pm0.08$ & $88.70\pm0.02$ & $93.83\pm0.11$ & 2.22 \\
Overall Top-1         & L2 & 95.10 & 93.73 & $92.11\pm0.10$ & $76.97\pm0.14$ & $91.27\pm0.11$ & 6.58 \\
Overall Top-1         & L3 & 95.10 & 91.79 & $88.38\pm0.26$ & $62.69\pm0.18$ & $87.35\pm0.15$ & 12.55 \\
\midrule
Medical Macro Recall  & L1 & 93.19 & 92.50 & $92.60\pm0.28$ & $85.80\pm0.32$ & $91.48\pm0.19$ & 2.59 \\
Medical Macro Recall  & L2 & 93.19 & 91.41 & $90.96\pm0.32$ & $74.14\pm0.73$ & $88.31\pm0.31$ & 6.98 \\
Medical Macro Recall  & L3 & 93.19 & 88.76 & $87.36\pm0.30$ & $60.28\pm0.42$ & $84.27\pm0.32$ & 13.02 \\
\bottomrule
\end{tabular}}
\end{table}

\section{Jetson Deployment Protocol}
\label{app:protocol}
\paragraph{Hardware and software environment.}
All deployment measurements were obtained on an NVIDIA Jetson Orin Nano Engineering Reference Developer Kit Super with an $8$\,GB-class module. The preserved software environment records Ubuntu 22.04.5 LTS, L4T R36.4.4 with kernel 5.15.148-tegra, TensorRT 10.3.0, CUDA 12.6, and cuDNN 9.3.0, and the benchmark artifacts are associated with the MAXN SUPER setup. The exact per-run \texttt{jetson\_clocks} state and cooling policy were not logged and are therefore not claimed.

\paragraph{TensorRT engine construction.}
Each model was exported to ONNX (opset 17, static input shape, constant folding) and its FP16 engine was built directly on the Jetson with \texttt{trtexec} using the common pattern \texttt{trtexec --onnx=$\langle$m$\rangle$ --saveEngine=$\langle$e$\rangle$ --fp16 --shapes=$\langle$s$\rangle$}. Workspace limits, builder optimization levels, tactic sources, and timing-cache files were left unspecified, so TensorRT defaults were used, and no DLA execution, INT8 calibration, sparsity option, CUDA Graph, or custom plugin was enabled. The same builder pattern was applied to every compared model while preserving each model's native input shape. Serialized engine size is computed as the plan-file byte size divided by $2^{20}$, so the reported engine sizes are in MiB.

\paragraph{Throughput measurement.}
Engines were benchmarked at batch size one with a $5000$\,ms warm-up and a $10$\,s timed window using one inference stream and \texttt{--useSpinWait}. \texttt{trtexec} feeds synthetic random input tensors, so real video decoding and dataset preprocessing are not part of the timing loop. Throughput is reported as clips per second, computed as $1000$ divided by the mean host latency, where one inference is one model-specific input clip ($16$ frames for Ours, the locally deployed EPAM-Net model, and DVANet; $40$ for SVdata2vec). The measured host-latency interval includes host-to-device transfer, TensorRT execution, and device-to-host transfer, and excludes video decoding, temporal sampling, person-centric cropping, resizing, normalization, file I/O, softmax, and top-1 postprocessing. Throughout, we report clip throughput in clips per second at batch size one rather than decoded video frames per second.

\paragraph{Memory and power measurement.}
System RAM and power were sampled with \texttt{tegrastats} at a requested interval of $100$\,ms. Peak RAM denotes the maximum absolute total system RAM observed during the benchmark session; it is not process RSS and is not idle-baseline-subtracted, and the monitoring window covers engine loading, context creation, warm-up, and timed inference, but not engine construction. Power denotes the mean \texttt{VDD\_IN} total-module input power over the monitored window and is likewise not idle-subtracted, so it should not be read as GPU-only power. TensorRT context memory is taken from the \texttt{trtexec} message ``Created execution context with device memory size'' and represents TensorRT-managed activation and scratch memory for one execution context, excluding serialized engine weights and model-independent CUDA runtime allocations. No thermal-throttling marker was observed in the preserved logs, with the recorded junction temperature remaining at or below approximately $63^{\circ}$C.

\paragraph{Scope of comparison.}
All models were evaluated on the same Jetson device with TensorRT 10.3, FP16 precision, batch size one, static input profiles, and a common \texttt{trtexec} procedure. Because the official EPAM-Net release provides only the full multimodal (RGB+pose) model and no standalone RGB-only checkpoint, the EPAM-Net entry here is our RGB-only re-implementation, trained under the same RGB-only protocol so that a comparable single-stream engine could be deployed. The workloads are nevertheless model-specific: Ours and the locally deployed EPAM-Net model use 16-frame RGB clips of shape $1\times3\times16\times224\times224$, DVANet uses a 16-frame RGB input of shape $1\times16\times3\times224\times224$, and SVdata2vec uses a 40-frame RGB input together with a skeleton input. Table~\ref{table3} should therefore be read as a deployment characterization of the preserved model configurations, not as a normalized per-frame or per-modality latency comparison. Each reported runtime value comes from a single benchmark session, and across-session variance was not recorded.

\begin{table}[t]
\centering
\caption{Reproducibility summary of the Jetson/TensorRT deployment protocol. No performance values are introduced in this table.}
\label{tab:protocol}
\footnotesize
\setlength{\tabcolsep}{4pt}
\begin{tabular}{@{}l p{5.1cm}@{}}
\toprule
Category & Setting \\
\midrule
Device & Jetson Orin Nano Eng.\ Ref.\ Dev.\ Kit Super, 8\,GB class \\
OS / L4T & Ubuntu 22.04.5 / L4T R36.4.4, kernel 5.15.148-tegra \\
Runtime & TensorRT 10.3.0, CUDA 12.6, cuDNN 9.3.0 \\
ONNX & Opset 17, static shape, constant folding \\
Precision & FP16 \\
Engine builder & \texttt{trtexec}, built on the Jetson device \\
Batch size & 1 \\
Warm-up & 5000\,ms \\
Timed window & 10\,s, single benchmark session \\
Streams & 1, \texttt{--useSpinWait} \\
Input source & \texttt{trtexec} synthetic random tensors \\
Timing boundary & H2D + TensorRT execution + D2H \\
Throughput unit & clips/s $=1000/$mean host latency \\
RAM & Peak absolute total system RAM (\texttt{tegrastats}); no idle subtraction \\
Power & Mean \texttt{VDD\_IN} total-module power; no idle subtraction \\
Context memory & \texttt{trtexec} execution-context device-memory field \\
\bottomrule
\end{tabular}
\end{table}

\section{Throughput Bottleneck Hypothesis}
\label{app:bottleneck}
We did not perform operator-level profiling (e.g., Nsight or \texttt{trtexec} traces) on the device, so we do not claim a specific bottleneck operator. A plausible explanation, consistent with the observed runtime behavior and with the structure of the graph, is the following. The architecture relies on multiple lightweight, memory-bound operators---TSM-style temporal shift with tensor reshaping, Ghost pointwise convolutions, $(2{+}1)$D factorized depthwise operators, and TAda-based temporal gating---which together tend to produce a fragmented operator graph with many small CUDA kernels. On embedded GPUs, such patterns are associated with increased kernel-launch overhead and CPU-side scheduling cost, while offering limited arithmetic intensity to fully utilize Tensor Cores. Moreover, TensorRT is most effective when it can fuse large, compute-dense operators; the narrow channel widths and frequent depthwise and reshape operations in our design are likely to reduce fusion opportunities and shift the bottleneck toward memory bandwidth and dispatch latency. We emphasize that this is a plausible, structure-consistent explanation rather than a profiled finding. Future work should verify the hypothesized bottlenecks through operator-level profiling and investigate hardware-aware alternatives for depthwise- and reshape-heavy operators.

\putbib[refs]
\end{bibunit}

\end{document}